\documentclass[11pt]{article}
\usepackage{geometry}
\usepackage{coling2020}
\usepackage{times}
\usepackage{url}
\usepackage{enumitem}
\usepackage{latexsym}
\usepackage{multirow}
\usepackage{hyperref}
\usepackage{microtype}
\usepackage{tcolorbox}
\usepackage{color,soul}
\usepackage[english]{babel}
\usepackage[utf8]{inputenc}
\usepackage{algorithm}
\usepackage{arevmath}     
\usepackage[noend]{algpseudocode}
\usepackage{mdframed}

\definecolor{airforceblue}{rgb}{0.36, 0.54, 0.66}
\definecolor{aurometalsaurus}{rgb}{0.43, 0.5, 0.5}
\definecolor{cadetgrey}{rgb}{0.57, 0.64, 0.69}
\definecolor{ceil}{rgb}{0.57, 0.63, 0.81}
\definecolor{gray(x11gray)}{rgb}{0.75, 0.75, 0.75}
\RequirePackage{colortbl}
\definecolor{airforceblue}{rgb}{0.36, 0.54, 0.66}
\definecolor{amaranth}{rgb}{0.9, 0.17, 0.31}
\definecolor{applegreen}{rgb}{0.55, 0.71, 0.0}
\definecolor{alizarin}{rgb}{0.82, 0.1, 0.26}
\definecolor{azure}{rgb}{0.0, 0.5, 1.0}
\definecolor{cadmiumgreen}{rgb}{0.0, 0.42, 0.24}

\colingfinalcopy 

\title{IIT Gandhinagar at SemEval-2020 Task 9: Code-Mixed Sentiment Classification Using Candidate Sentence Generation and Selection}

\author{{Vivek Srivastava, Mayank Singh} \\ {Indian Institute of Technology Gandhinagar} \\ {Gujarat India} \\ \texttt{\{vivek.srivastava, singh.mayank\}@iitgn.ac.in}}
\begin{document}
\maketitle
\begin{abstract}
Code-mixing is the phenomenon of using multiple languages in the same utterance of a text or speech. It is a frequently used pattern of communication on various platforms such as social media sites, online gaming, product reviews, etc. Sentiment analysis of the monolingual text is a well-studied task. Code-mixing adds to the challenge of analyzing the sentiment of the text due to the non-standard writing style. We present a candidate sentence generation and selection based approach on top of the Bi-LSTM based neural classifier to classify the Hinglish code-mixed text into one of the three sentiment classes positive, negative, or neutral. The proposed approach shows an improvement in the system performance as compared to the Bi-LSTM based neural classifier. The results present an opportunity to understand various other nuances of code-mixing in the textual data, such as humor-detection, intent classification, etc.
  
\end{abstract}

\section{Introduction}
\label{intro}
 Code mixing is one of the most frequent styles of communication in multilingual communities, such as India. This pattern of communication on various platforms such as social media, online gaming, online product reviews, etc. makes it difficult to understand the sentiment of the text. Sentiment classification of the code-mixed text is useful in the scenarios of socially or politically driven discussions, fake news propagation,  etc. Some of the major challenges with the text in the code-mixed language are:
    
    \begin{itemize}[noitemsep,nolistsep]
    \item\textbf{Ambiguity in language identification}:  \textit{is, me, to} are some examples of the words that are ambiguous to classify as English and Hindi without proper knowledge of context.
    
    \item \textbf{Spelling variations}: E.g., \textit{jaldi, jldi, jldiii,..} are some variations for the word \textit{hurry} in English.
    
    \item \textbf{Misplaced/ skipped punctuation}: E.g., \textit{Aap kb se cricket khelne lage..never saw u bfr}. \newline The sentence in the example misses a question mark(?) apart from other necessary modifications to make the structure of the sentence correct. 
    
    \item \textbf{Missing context}:  E.g., \textit{Note kr lijiye.. Bandi chal rahi h ;)} is a code-mixed sentence and demonetisation (notebandi) is the hidden context.
    \end{itemize}
   
    With the increasing popularity of using code-mixing on social media platforms, the interest to study the various dynamics of code-mixing is seeking a boom. Multiple works on language identification \cite{barman-etal-2014-code,das2014identifying}, POS tagging \cite{vyas2014pos,ghosh2016part}, named entity recognition \cite{singh2018language,singh2018named}, etc. shows the challenges and the opportunities with the code-mixed data. \newcite{pang2008opinion} presents a survey of the approaches to understand the opinions and sentiments on various platforms. \newcite{dos2014deep} performs the sentiment analysis task of the short text messages on two corpora from different domains and present their findings. \newcite{kouloumpis2011twitter} presents multiple experiments to understand the sentiment of Twitter messages using linguistic features and lexical resources. Sentiment analysis of the code-mixed Tweets using a sub-word level representation \cite{prabhu2016subword} in the LSTM can improve the performance of the system. \newcite{swami2018corpus} presents a corpus to understand and detect the sarcasm of 5250 code-mixed English-Hindi tweets.
    
     \textbf{Contributions}: We present a candidate sentence generation and selection based procedure on top of the Bi-LSTM neural classifier. We observe the increase in the system performance by using the proposed architecture as compared to the Bi-LSTM classifier.

\section{Dataset}
\label{dataset}
We use the dataset \cite{patwa2020sentimix} provided by the task organizers for building our system (Codalab username: vivek\_IITGN). Each sentence in the dataset has a sentiment label as positive, negative, or neutral. Table \ref{tab: ditribution} shows the distribution of the sentences in the train, validation, and test dataset for each class. We have 15131, 3000, and 3000 sentences in train, validation, and test set, respectively. On manual inspection of the dataset, we observe ambiguity in the annotation of the sentences. To examine this further, we extract the top 20 most frequently used words in the dataset. We remove the English stopwords, and we set a threshold of 4 characters on the length of the tokens to filter out the Romanized Hindi stopwords. Table \ref{tab:overlap} shows the percentage overlap of most frequent 20 words of length more than four characters in the train, validation, and test set. The high percentage overlap of most frequent neutral words with positive and negative words also indicates the presence of ambiguity as a challenge in the annotation. Ambiguity in the label for the sentence is one of the major challenges for understanding the sentiment of the sentence. Figure \ref{fig:ambiguous} shows some of the example sentences in training set with ambiguous sentiment label. There could be multiple reasons for the ambiguity in the annotation of the sentences such as hidden sarcasm, targeting individual or institution, unclear intent, etc. It leads to human bias due to the annotator's perception of the event or the individual in the sentence.
To preprocess the dataset, we remove hyperlinks, mentions, hashtags, emoticons, and special characters from the sentences. We lowercase the sentences. To identify and remove the emoticons from the sentences, we use the emoji sentiment dataset\footnote{https://www.kaggle.com/thomasseleck/emoji-sentiment-data}. 

\begin{table}[h]
\centering
\begin{tabular}{|c|c|c|l|}
\hline
                            & \textbf{Train} & \textbf{Validation} & \textbf{Test} \\ \hline
\textbf{Positive Sentences} & 5034           & 982                 & 1000          \\ \hline
\textbf{Negative Sentences} & 4459           & 890                 & 900           \\ \hline
\textbf{Neutral Sentences}  & 5638           & 1128                & 1100          \\ \hline
\end{tabular}
\caption{Distribution of the sentences in the train, validation, and test set.}
\label{tab: ditribution}
\end{table}

\begin{table}[h]
\resizebox{\hsize}{!}{
\begin{tabular}{|c|c|c|c||c|c|c||c|c|c|}
\hline
                  & \multicolumn{3}{c|}{\textbf{Training Set}}               & \multicolumn{3}{c|}{\textbf{Validation Set}}             & \multicolumn{3}{c|}{\textbf{Test Set}}                   \\ \hline
                  & \textbf{Positive} & \textbf{Negative} & \textbf{Neutral} & \textbf{Positive} & \textbf{Negative} & \textbf{Neutral} & \textbf{Positive} & \textbf{Negative} & \textbf{Neutral} \\ \hline
\textbf{Positive} & -                 & 20                & 55               & -                 & 20                & 50               & -                 & 15                & 40               \\ \hline
\textbf{Negative} & 20                & -                 & 55               & 20                & -                 & 45               & 15                & -                 & 40               \\ \hline
\textbf{Neutral}  & 55                & 55                & -                & 50                & 45                & -                & 40                & 40                & -                \\ \hline
\end{tabular}
}
\caption{Percentage overlap of the most frequent 20 words of length more than 4 characters in the training, validation, and test set.}
\label{tab:overlap}
\end{table}

\begin{figure}[h]
    \centering
    \begin{tcolorbox}[colback=white]

\textsc{Code-mixed Sentence}: \textcolor{alizarin}{Twitter k baghair apna roza mumkin nahi hota ? Apna chutiyaap dusron per thopna band karo Bhai ! https // t . co / APKD4G8lh0}\\
\textsc{Original label}: \textcolor{cadmiumgreen}{Positive}

\textsc{Code-mixed Sentence}: \textcolor{alizarin}{@ JDeepDhillonz Ha ha ha isko issi baat ka darr the tabhi Congi se alliance ke peechey pada hua tha !}\\
\textsc{Original label}: \textcolor{cadmiumgreen}{Negative}

\textsc{Code-mixed Sentence}: \textcolor{alizarin}{@ Shaan \_ pathan \_ 14 @ DwivediAnukriti Ikk toh sarkar job ni de rhi or upper se apne india ke log kaam karna nahi chahat … https // t . co / zfkm4obLd6}\\
\textsc{Original label}: \textcolor{cadmiumgreen}{Neutral}

\end{tcolorbox}
\caption{Example sentences from the training set with ambiguous labels.}
\label{fig:ambiguous}
\end{figure}

\section{Experiments}
\label{experiment}

 The availability of code-mixed embedding is a challenging task due to the scarcity of large scale code-mixed corpora. We are using the Glove embedding \cite{pennington2014glove} for the English words, and we train the embedding on the PHINC dataset \cite{srivastava2020phinc} for the Romanized Hindi words. We are using the code-mixed sentences from PHINC to train the code-mixed embedding.

\begin{figure}[h]
\centering
\includegraphics[width=1\linewidth]{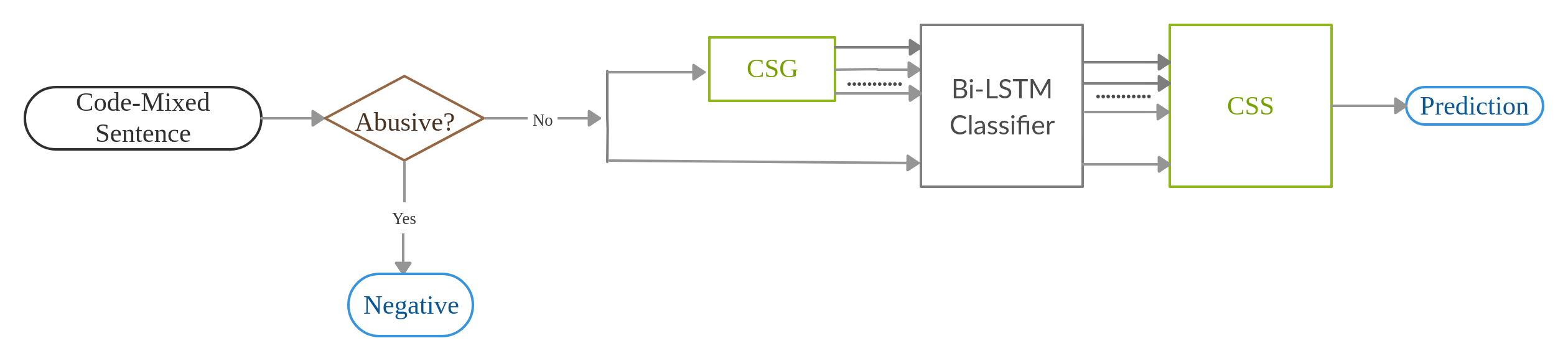}
\caption{Flow diagram of the proposed approach.}
\label{fig:flow}
\end{figure}

Initially, we train the system using Bi-LSTM based neural architecture. The architecture of the Bi-LSTM classifier has the embedding layer followed by the Bi-LSTM layer and then two dense layers and, finally, the softmax prediction for the three sentiment classes. For prediction on the test set, we pre-filter the sentence based on the list of abusive words. If a sentence contains any words from this list, we label that sentence as negative. In the pre-filtering process, we identify 123 sentences in the test set containing one or more of the abusive words from the list. Post pre-filtering step, we generate 15 candidate sentences for each of the remaining test instance using the Candidate Sentence Generation (CSG) procedure. We then select the best sentiment prediction for the sentence using the Candidate Sentence Selection (CSS) procedure. Algorithm 1 shows the CSG procedure. Algorithm 2 shows the CSS procedure. Figure \ref{fig:flow} shows the flow diagram of the proposed approach. In the CSG procedure, we try to confuse the model with nearly similar sentences with additional phrases. We generate five similar sentences to the original code-mixed sentence for each of the three buckets (positive, negative, and neutral). We detect the degree of confusion in the sentiment prediction using the CSS procedure. We also keep track of the degree of confusion by sentences in each bucket. If the degree of confusion is significantly higher, we change the previous prediction by the model using the rules (as discussed in Algorithm 2).

\begin{algorithm}
\caption{Candidate Sentence Generation (CSG) procedure}
\begin{mdframed}[backgroundcolor=gray!3]
\begin{algorithmic}[1]

\Procedure{CSG}{$CM\_sent_i$}      
    \State Load the set of positive and negative phrases P. Table \ref{tab:phrases} shows the set P.
   \State Set $bucket_1$=[], $bucket_2$=[], and $bucket_3$=[]
   \For{i=1$\rightarrow$5}
   \State Create a set $spots_{pos}$=$\{sp_1, sp_2,...\}$ of randomly selected spots in $CM\_sent_i$
   \State Create a set $spots_{neg}$=$\{sn_1, sn_2,...\}$ of randomly selected spots in $CM\_sent_i$
   \State Create a set $spots_{neu}$=$\{su_1, su_2,...\}$ of randomly selected spots in $CM\_sent_i$
   \State Set $sent_{pos}$=$CM\_sent_i$, $sent_{neg}$=$CM\_sent_i$, and $sent_{neu}$=$CM\_sent_i$
   \For{each spot $sp_j$ in $spots_{pos}$}
   \State Randomly select a positive phrase $p_{pos}$ from the set P
   \State Replace the phrase $p_{pos}$ at spot $sp_j$ in $sent_{pos}$
   \EndFor
   \State Add the new sentence $sent_{pos}$ in the list $bucket_1$
   \For{each spot $sn_j$ in $spots_{neg}$}
   \State Randomly select a negative phrase $p_{neg}$ from the set P
   \State Replace the phrase $p_{neg}$ at spot $sn_j$ in $sent_{neg}$
   \EndFor
   \State Add the new sentence $sent_{neg}$ in the list $bucket_2$
   \For{each spot $su_j$ in $spots_{neu}$}
   \State Alternatively select phrase $p_{pos}$ and $p_{neg}$ from the set P
   \State  Replace alternatively the phrase $p_{pos}$ and $p_{neg}$ at spot $su_j$ in $sent_{neu}$
   \EndFor
   \State Add the new sentence $sent_{neu}$ in the list $bucket_3$
   \EndFor
   \State Return $bucket_1$, $bucket_2$, and $bucket_3$
\EndProcedure

\end{algorithmic}
\end{mdframed}
\end{algorithm}

\begin{table}[!tbh]
\centering
\resizebox{\hsize}{!}{
\begin{tabular}{|c|c|}
\hline
                  & \textbf{Phrases}                                                                                             \\ \hline
\textbf{Positive} & \begin{tabular}[c]{@{}c@{}}good enough, sure thing, undoubtedly, theek hai, even so, indubitably, tathaastu, of course, savaida, certainly, \\ sakaaraatmak, gladly, affirmative, apanee marjee, abhee to, sabase adhik aashvast, amen, good, yakeenan, bahut achchha,\\  theek, precisely, by all means, beyond a doubt, surely, yeah, unquestionably, very well, exactly, positively, khushee se, \\ har tarah se, most assuredly, definitely, achchha\end{tabular} \\ \hline
\textbf{Negative} & \begin{tabular}[c]{@{}c@{}}nowhere, nahin kar sakate, koee bhee nahin, nahin, wouldn’t, won’t, nahin hai, nahin karana chaahie, nahin kiya ja saka,\\  kabhee naheen, never, don’t, neither, couldn’t, nothing, doesn’t, koee nahin, barely, mushkil se, kuchh bhee to nahin, wasn’t, \\ shouldn’t, scarcely,  nahin karata hai, hardly, nahin hoga, kaheen bhee nahin, no, not, na, nobody, can’t, shaayad hee, no one, none\end{tabular}                            \\ \hline
\end{tabular}
}
\caption{List of positive and negative phrases.}
\label{tab:phrases}
\end{table}

\begin{figure}[!tbh]
\centering
\begin{tcolorbox}[colback=white]

\textsc{Code-mixed pre-processed sentence}: \textcolor{alizarin}{rohit bhai i am your big fan want to meet you} \textcolor{alizarin}{}\\
\textsc{Sentence with spots (marked as $<$spot$>$)}: \textcolor{blue}{rohit bhai i am $<$spot$>$ your big fan $<$spot$>$ want to meet you}\\
\textsc{Sentence in Bucket 1}: \textcolor{cadmiumgreen}{rohit bhai i am \textit{undoubtedly} your big fan \textit{by all means} want to meet you}\\
\textsc{Sentence in Bucket 2}: \textcolor{cadmiumgreen}{rohit bhai i am \textit{scarcely} your big fan \textit{shaayad hee} want to meet you}\\
\textsc{Sentence in Bucket 3}: \textcolor{cadmiumgreen}{rohit bhai i am \textit{positively} your big fan \textit{kaheen bhee nahin} want to meet you}

\end{tcolorbox}
\caption{Example CSG procedure for sentences in all the three buckets. We select phrases in \textit{italics} at random from the list of positive and negative phrases and replace with $<$spot$>$ as per the rule for each bucket.}
\label{fig:CSG}
\end{figure}

\begin{algorithm}
\caption{Candidate Sentence Selection (CSS) procedure}
\begin{mdframed}[backgroundcolor=gray!3]
\begin{algorithmic}[1]

\Procedure{CSG}{$pred\_sent_i$, $pred\_bucket_1$, $pred\_bucket_2$,$pred\_bucket_3$}      
\State Set $pred_{final}$=[] 
\If{$pred\_sent_i$ is \textit{Positive}}
\If{most frequent prediction in $pred\_bucket_1$ is \textit{Positive}}
\State Set $pred_{final}$= \textit{Positive}
\Else 
\State Set $pred_{final}$= most frequent prediction in $pred\_bucket_2$ and $pred\_bucket_3$
\EndIf

\ElsIf{$pred\_sent_i$ is \textit{Negative}}
\If{most frequent prediction in $pred\_bucket_2$ is \textit{Negative}}
\State Set $pred_{final}$= \textit{Negative}
\Else 
\State Set $pred_{final}$= most frequent prediction in $pred\_bucket_1$ and $pred\_bucket_3$
\EndIf

\Else
\If{most frequent prediction in $pred\_bucket_3$ is \textit{Neutral}}
\State Set $pred_{final}$= \textit{Neutral}
\Else 
\State Set $pred_{final}$= most frequent prediction in $pred\_bucket_1$ and $pred\_bucket_2$
\EndIf
\EndIf
\State Return $pred_{final}$
\EndProcedure
\end{algorithmic}
\end{mdframed}
\end{algorithm}

\section{Results and Analysis}
\label{result}
To evaluate the system performance, we use accuracy, precision, recall, and f-score as the evaluation metric. Table \ref{tab:CSS} shows the distribution of the successfully and unsuccessfully modified sentences for the final prediction by the Bi-LSTM + CSG + CSS model. We use the prediction by the Bi-LSTM classifier as the baseline. We observe relatively better successful modifications for the neutral sentences to and from the positive and negative sentences. This result can be attributed to the high overlap in the most frequent words in the neutral sentences with both the other classes (as discussed in section \ref{dataset}). 
Table \ref{tab: results} shows the system performance of the two models on the test dataset. We observe an increase in the system performance with the use of CSG and CSS procedures on top of the Bi-LSTM classifier. Table \ref{tab:class_res} shows the system performance on the test dataset with the classwise F-score as the evaluation metric. 

\begin{table}[h]
\centering
\begin{tabular}{|c|c|c|c||c|c|c|}
\hline
\multirow{2}{*}{} & \multicolumn{3}{c|}{\textbf{Successful modification}}          & \multicolumn{3}{c|}{\textbf{Unsuccessful modification}}        \\ \cline{2-7} 
                  & \textbf{Positive} & \textbf{Negative} & \textbf{Neutral} & \textbf{Positive} & \textbf{Negative} & \textbf{Neutral} \\ \hline
\textbf{Positive} & -                 & 10                & 77               & -                 & 20                & 96               \\ \hline
\textbf{Negative} & 7                 & -                 & 120              & 34                & -                 & 149              \\ \hline
\textbf{Neutral}  & 189               & 95                & -                & 153               & 100               & -                \\ \hline
\end{tabular}
\caption{Distribution of the successful and unsuccessful modification of the test sentences by the Bi-LSTM + CSG + CSS model. We use Bi-LSTM classifier as the baseline. Prediction labels in the rows shows the prediction by the Bi-LSTM classifier whereas the prediction labels in the columns are for the Bi-LSTM + CSG + CSS model.}
\label{tab:CSS}
\end{table}

\begin{table}[!tbh]
\centering
\begin{tabular}{|c|c|c|c|c|}
\hline
                            & \textbf{Accuracy} & \textbf{Precision} & \textbf{Recall} & \textbf{F-score} \\ \hline
\textbf{Bi-LSTM}            & 0.587             & 0.607              & 0.590           & 0.595            \\ \hline
\textbf{Bi-LSTM + CSG + CSS} & 0.608             & 0.618              & 0.609           & 0.612            \\ \hline
\end{tabular}
\caption{System performance on the test dataset. We use the macro score for evaluation.}
\label{tab: results}
\end{table}

\begin{table}[!tbh]
\centering
\begin{tabular}{|c|c|c|c|}
\hline
                             & \textbf{Positive} & \textbf{Negative} & \textbf{Neutral} \\ \hline
\textbf{Bi-LSTM}             & 0.621             & 0.643             & 0.520            \\ \hline
\textbf{Bi-LSTM + CSG + CSS} & 0.754             & 0.625             & 0.459            \\ \hline
\end{tabular}
\caption{Evaluation of the system performance based on classwise F-score.}
\label{tab:class_res}
\end{table}

\section{Conclusion and Future Work}
We present a Bi-LSTM based sentiment classifier for the classification of code-mixed Hinglish sentences. We also propose a candidate sentence generation and selection based approach on top of the Bi-LSTM based classifier to improve the system performance. Up to a certain extent, the proposed approach is able to detect the ambiguous labels in the dataset. We can extend the proposed method to solve other challenges relevant to code-mixing, such as sarcasm detection, fake-news identification, intent classification, etc. 

\clearpage
\bibliographystyle{coling}
\bibliography{semeval2020}

\end{document}